\documentclass[conference]{IEEEtran}

\usepackage{amsmath,amssymb}
\usepackage{bm}
\usepackage{graphicx}
\usepackage{booktabs}
\usepackage{multirow}
\usepackage{siunitx}
\usepackage{xcolor}
\usepackage{url}
\usepackage{balance}
\usepackage[capitalise]{cleveref}
\crefname{equation}{}{}
\crefname{figure}{Fig.}{Figs.}
\crefname{tabular}{Tab.}{Tabs.}
\crefname{section}{Sec.}{Secs.}
\usepackage{xspace}

\title{The Talking Robot: Distortion-Robust Acoustic Models for Robot-Robot Communication}

\author{
  \IEEEauthorblockN{Hanlong Li\IEEEauthorrefmark{1}, Karishma Kamalahasan\IEEEauthorrefmark{2}, Jiahui Li\IEEEauthorrefmark{1}, Kazuhiro Nakadai\IEEEauthorrefmark{1}, and Shreyas Kousik\IEEEauthorrefmark{2}}
  \IEEEauthorblockA{\IEEEauthorrefmark{1}Institute of Science Tokyo\\
  \{li.h.5ead@m.isct.ac.jp, li.j.68a5@m.isct.ac.jp, nakadai@ra.sc.e.titech.ac.jp\}}
  \IEEEauthorblockA{\IEEEauthorrefmark{2}Georgia Institute of Technology\\
  \{kkamalahasan3@gatech.edu, shreyas.kousik@me.gatech.edu\}}
}

\newcommand{\ttsmel}{TTS\_MEL\xspace}
\newcommand{\ttswav}{TTS\_WAV\xspace}
\newcommand{\pswav}{PS\_WAV\xspace}

\begin{document}
\maketitle

\begin{abstract}
We present Artoo, a learned acoustic communication system for robots that
replaces hand-designed signal processing with end-to-end co-trained
neural networks. Our system pairs a lightweight text-to-speech (TTS)
transmitter (1.18M parameters) with a conformer-based
automatic speech recognition (ASR) receiver (938K parameters),
jointly optimized through a differentiable channel.
Unlike human speech, robot-to-robot communication is
\emph{paralinguistics-free}: the system need not preserve timbre,
prosody, or naturalness, only maximize decoding accuracy under
channel distortion. Through a three-phase co-training curriculum, the TTS transmitter
learns to produce distortion-robust acoustic encodings that
surpass the baseline under noise, achieving
\textbf{8.3\% CER} at 0~dB SNR. The entire system requires only
2.1M parameters (8.4~MB) and runs in under
13~ms end-to-end on a CPU, making it suitable for
deployment on resource-constrained robotic platforms.
\end{abstract}

\begin{IEEEkeywords}
robot communication, speech chain, co-training, conformer, FastSpeech2, edge inference, noise robustness
\end{IEEEkeywords}

\section{Introduction}
Robots that share a workspace often need to exchange short commands quickly and reliably. While radio-frequency links are effective, they can introduce practical overhead (dedicated transceivers, interference management, and deployment constraints). Acoustic communication is a lightweight alternative for short-range, infrastructure-free messaging.

Our key observation is that robot-to-robot acoustic communication is \emph{paralinguistics-free}: the signal does not need to resemble human speech or preserve voice characteristics. The only objective is to recover a discrete token sequence. This lets us treat the transmitter and receiver as an encoder--decoder pair connected through an acoustic channel, in the spirit of learned communication systems~\cite{oshea2017introduction}.

We begin with a deterministic \textit{procedural synthesizer} (PS) that assigns each of the 128 vocabulary tokens a unique multi-harmonic tone chip. PS provides full vocabulary coverage at zero data cost and an interpretable baseline that a compact receiver can decode with near-zero error in clean conditions. However, PS is brittle under realistic distortions: reverberation smears token boundaries, clipping injects overlapping harmonics, and sample-rate drift shifts frequencies enough to cause collisions. These failures motivate replacing fixed, hand-designed signaling with end-to-end learning.

We propose co-training a TTS-style transmitter and a Conformer-based receiver through realistic channel augmentation. The transmitter (FastSpeech~\cite{ren2019fastspeech}) maps tokens to mel-spectrograms as a learned acoustic code; a Griffin--Lim vocoder~\cite{griffin1984signal} converts the spectrogram to a waveform for physical transmission; the receiver (Conformer~\cite{gulati2020conformer}) decodes tokens via CTC~\cite{graves2006ctc}. Direct end-to-end training from scratch fails due to the \emph{cold-start problem}. We resolve this with a three-phase curriculum that uses PS as a warm-start anchor: (i) warmup (ASR learns PS; TTS imitates PS), (ii) ramp (gradually enable CTC$\rightarrow$TTS), and (iii) full co-training (remove PS; train end-to-end under augmentation). The resulting system matches PS in clean conditions and surpasses it under degradation, while remaining lightweight (\SI{2.1}{M} parameters) and real-time.

\begin{figure}[t]
    \centering
    \includegraphics[width=0.99\columnwidth]{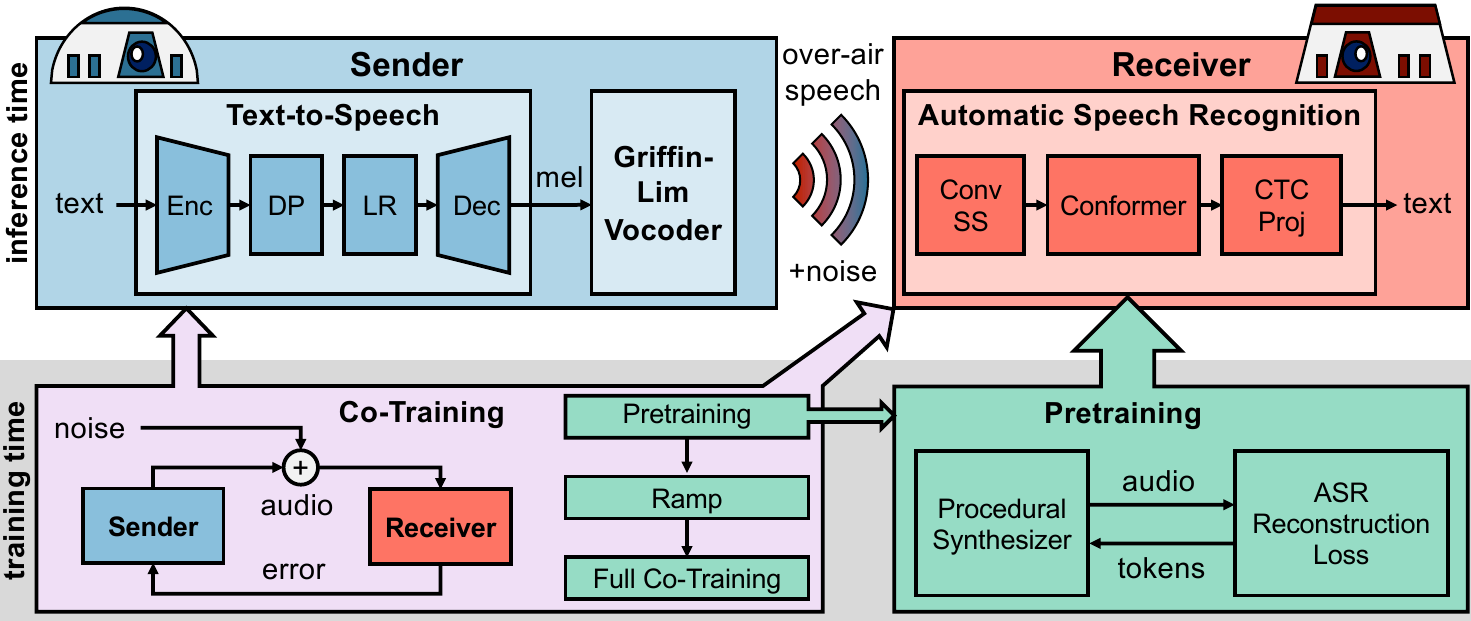}
    \caption{We propose \textit{Artoo}: a text-to-speech (TTS) and automatic speech recognition (ASR) pipeline for robot-robot communication.
    The goal of our method is to enable robots to communicate in a way that is robust to environmental noise; our key insight is to relax the need for TTS and ASR models to replicate human speech.}
    \label{fig: front figure}
    \vspace*{-10pt}
\end{figure}

\subsection{Contributions}
\begin{itemize}
\item We cast robot acoustic messaging as paralinguistics-free discrete symbol transmission and repurpose compact TTS/ASR models as a learned acoustic codec, called \textit{Artoo} (\underline{a}coustic \underline{r}obot \underline{t}ransmission r\underline{o}bust c\underline{o}dec).
\item We introduce a procedural synthesizer baseline that provides a zero-data proof of concept and resolves cold-start as a curriculum anchor for stable co-training.
\item We demonstrate sim-to-real robustness via end-to-end co-training under realistic channel augmentation, achieving reliable over-the-air decoding with real-time latency on embedded hardware.
\end{itemize}

\subsection{Paper Organization}
Next, we present related work in \Cref{sec:relatedwork}.
We propose our Artoo system architecture in \Cref{sec:arch}, then explain how we train it in \Cref{sec:method}.
We set up our experiments in \Cref{sec:exp}, present them in \Cref{sec:results}, and discuss advantages and limitations of our approach in \Cref{sec:discussion}.
We provide closing remarks in \Cref{sec:conclusion}.
\section{Related Work}\label{sec:relatedwork}
\subsection{Acoustic Data Communication}

Acoustic data links have been explored for short-range
device-to-device communication. GGWave~\cite{gerganov2020ggwave}
(also deployed as Gibberlink) uses multi-frequency FSK with
Reed-Solomon error correction, achieving reliable data transfer
over speaker-microphone channels. Chirp~\cite{chirp2016} encodes
data into audible or near-ultrasonic chirp signals. These systems
rely on hand-crafted modulation, synchronization preambles, and
forward error correction (FEC) designed by signal processing
experts. While robust within their design parameters, they cannot
adapt to novel channel conditions without manual re-engineering.

Underwater acoustic communication has driven significant work on
learned modulation. Li et al.~\cite{li2020deep} apply deep
learning to underwater channel equalization, and recent work
explores end-to-end autoencoder-based underwater acoustic
modems~\cite{underwater2023}. Our work targets the less-explored
in-air acoustic channel between co-located robots.

\subsection{Learned Communication Systems}

O'Shea and Hoydis~\cite{oshea2017introduction} proposed treating
an entire communication system as an autoencoder, where the
transmitter (encoder), channel, and receiver (decoder) are
jointly optimized. This framework has been extended to OFDM
systems~\cite{felix2018ofdm}, MIMO channels~\cite{oshea2017mimo},
and physical-layer security~\cite{fritschek2019deep}. However,
these works operate on baseband I/Q samples or constellation
diagrams in the RF domain. Our system operates in the acoustic
domain and uses speech-native architectures (TTS and ASR) as
the encoder and decoder, inheriting their sequence-modeling
capabilities for variable-length token streams.

\subsection{Neural Text-to-Speech and ASR}

Modern TTS systems such as FastSpeech 2~\cite{ren2020fastspeech2},
VITS~\cite{kim2021vits}, and Tacotron 2~\cite{shen2018natural}
produce high-quality mel-spectrograms from text. Conformer-based
ASR~\cite{gulati2020conformer} with CTC decoding~\cite{graves2006ctc}
achieves state-of-the-art recognition accuracy. We repurpose
these architectures not for human speech but as encoder-decoder
components of a communication system, stripping away modules
for naturalness (e.g., variance adaptors for pitch and energy)
and adding a self-consistency loss through Griffin-Lim
vocoding~\cite{griffin1984signal}.

\section{Artoo System Architecture}
\label{sec:arch}

We propose \textit{Artoo}, a co-trained text-to-speech (TTS) and automatic speech recognition (ASR) system for acoustic robot communication.
We explain how we train this system's parameters later in \Cref{sec:method}.

\subsection{System Overview}
\subsubsection{Artoo Deployment Pipeline}
A TTS module takes in a token sequence and outputs a mel-spectrogram.
This is fed into a Griffin-Lim Vocoder to generate an audio waveform.
The audio is transmitted over the air to an ASR module that performs mel extraction and speech-to-text.
The complete system comprises approximately 2.1M parameters, making it suitable for resource-constrained robotic platforms.

\subsubsection{Vocabulary/Tokens}
The system maps a token sequence $\bm{x} = (x_1, \ldots, x_L)$
from a fixed vocabulary $\mathcal{V}$ ($|\mathcal{V}|=128$) to an
acoustic waveform, transmits it over an in-air channel, and decodes
it back to a token sequence $\hat{\bm{x}}$.
The vocabulary covers 26 lowercase letters, 10 digits, 16 punctuation marks, 6 special tokens (blank, pad, SOS, EOS, space, UNK), and 44 task-specific robot commands (e.g., \texttt{<STOP>}, \texttt{<ACK>}, \texttt{<SCAN>}).

\subsubsection{Pretraining Architecture}
We use a Procedural Synthesizer (PS) to generate unique waveforms for each token, taking advantage of the fact that we do not need to replicate human speech; we find pretraining with these tokens improves ASR robustness.
Details are below in \Cref{sec:ps}.

\subsection{TTS Transmitter}
\label{sec:tts}
The TTS transmitter follows a non-autoregressive FastSpeech-style~\cite{ren2019fastspeech} architecture with $\sim$1.18M parameters (Table~\ref{tab:arch}) and and consists of four components:

\begin{enumerate}
    \item \textbf{Transformer Encoder}: A token embedding layer ($V \times d$) followed by sinusoidal positional encoding and $L_{\text{enc}}$ Transformer encoder layers with GELU activation.
    \item \textbf{Duration Predictor}: A two-layer 1D convolutional network with batch normalization that predicts log-duration for each input token:
    \begin{equation}
        \hat{d}_i = \text{round}\!\left(\exp(\text{DurPred}(\mathbf{h}_i)) - 1\right), \quad \hat{d}_i \geq 1
    \end{equation}
    where $\mathbf{h}_i$ is the encoder hidden state for token $i$.
    \item \textbf{Length Regulator}: Expands the token-level representations to mel-frame-level by repeating each hidden state $\mathbf{h}_i$ according to its predicted duration $\hat{d}_i$.
    \item \textbf{Transformer Decoder}: $L_{\text{dec}}$ Transformer encoder layers, followed by a linear projection to $N_{\text{mel}}$ dimensions.
\end{enumerate}

The output is a mel-spectrogram $\mathbf{M} \in \mathbb{R}^{N_{\text{mel}} \times T_{\text{mel}}}$. Crucially, the TTS does not include a vocoder, but rather outputs mel-spectrograms, not waveforms. This design enables direct gradient flow from the ASR loss back through the TTS during training.

Unlike a standard TTS, we removed pitch and energy variance adapters. The main reason is the transmitter does not need to produce natural-sounding speech. The only goal is to produce mel patterns that the ASR receiver can decode under channel corruption

\textbf{\begin{table}[t]
    \centering
    \caption{TTS hyperparameters.}
    \label{tab:tts_params}
    \begin{tabular}{lc}
        \toprule
        Parameter & Value \\
        \midrule
        Model dimension $d$ & 128 \\
        Feed-forward dimension $d_{\text{ff}}$ & 256 \\
        Attention heads & 4 \\
        Encoder layers $L_{\text{enc}}$ & 4 \\
        Decoder layers $L_{\text{dec}}$ & 4 \\
        Mel bins $N_{\text{mel}}$ & 40 \\
        Max mel length $T_{\text{mel}}$ & 300 \\
        \midrule
        Total parameters & $\sim$1.18M \\
        \bottomrule
    \end{tabular}
\end{table}}

\subsection{Griffin-Lim Vocoder}
\label{sec:gl}

To bridge TTS mel-spectrograms to physical waveforms, we use
the Griffin-Lim algorithm~\cite{griffin1984signal} with 32
iterations. Given the TTS output $\bm{M}$, we compute:
\begin{equation}
\hat{w} = \text{GL}\bigl(\text{pinv}(\bm{F}_\text{mel})
          \cdot \sqrt{\exp(\bm{M}) - 1}\bigr)
\label{eq:gl}
\end{equation}
where $\bm{F}_\text{mel}$ is the mel filterbank and
$\text{pinv}(\cdot)$ its pseudo-inverse. Griffin-Lim has
\textbf{zero learned parameters}, which we make as a deliberate choice.
Any neural vocoder (e.g., HiFi-GAN~\cite{kong2020hifi}) would
improve waveform quality, but Griffin-Lim establishes a
worst-case lower bound: if the system works with GL, it will
work at least as well with a better vocoder.

\subsection{ASR Receiver}
\label{sec:asr}

The ASR receiver is a Conformer-Tiny~\cite{gulati2020conformer} architecture with CTC output, totaling $\sim$938K parameters.

\begin{enumerate}
\item \textbf{Convolutional Subsampling.} Two strided Conv2D
      layers ($1 \to 32 \to 32$ channels, stride 2, GELU)
      reduce the time axis by $4\times$, followed by a linear
      projection to $d_\text{model}=128$.
\item \textbf{Conformer Blocks.} Four blocks, each comprising:
      half-step FFN $\to$ multi-head self-attention (4 heads)
      $\to$ depthwise separable convolution ($k\!=\!15$) $\to$
      half-step FFN $\to$ LayerNorm.
\item \textbf{CTC Projection.} A linear layer maps to
      $|\mathcal{V}|=128$ logits. Decoding uses greedy
      CTC collapse (remove blanks and repeated tokens).
\end{enumerate}

\subsection{Procedural Synthesizer}
\label{sec:ps}

The procedural synthesizer (PS) generates a deterministic
waveform for any token sequence without learned parameters.
Each token $x_i$ is assigned a unique 3-harmonic signature:
\begin{equation}
s_i(t) = \sum_{k=1}^{3} a_k \sin(2\pi f_k t + \varphi_k)
         \cdot w(t)
\label{eq:ps}
\end{equation}
where $f_k$, $a_k$, $\varphi_k$ are the frequency, amplitude,
and phase of the $k$-th harmonic, and $w(t)$ is a 5\,ms
fade-in/out window. Each token occupies a 60\,ms chip
(960 samples at 16\,kHz). Robot command tokens use golden-ratio
frequency spacing to maximize spectral separation:
$f_1 = 300 + (i \cdot \phi \cdot 83) \bmod 3500$\,Hz,
where $\phi = (1+\sqrt{5})/2$.

The PS is \textbf{not} a communication protocol, as it has no synchronization preamble, no framing, and no error correction.
Its sole purpose is to provide a \emph{physically realizable} mapping from tokens to waveforms that covers the full 128-token vocabulary at zero data cost, bootstrapping the ASR receiver before co-training begins.

\begin{table}[t]
\centering
\caption{Artoo Model architecture summary.}
\label{tab:arch}
\begin{tabular}{@{}lcc@{}}
\toprule
\textbf{Component} & \textbf{Params} & \textbf{Key Details} \\
\midrule
TTS Transmitter & 1.18\,M &
  \begin{tabular}[c]{@{}c@{}}4+4 Transformer layers,\\
  $d$=128, 4 heads\end{tabular} \\
\addlinespace
Griffin-Lim & 0 & 32 iterations \\
\addlinespace
ASR Receiver & 938\,K &
  \begin{tabular}[c]{@{}c@{}}4 Conformer blocks,\\
  Conv $k$=15, CTC\end{tabular} \\
\addlinespace
Procedural Synth. & 0 & 3-harmonic, 60\,ms/token \\
\midrule
\textbf{Total} & \textbf{2.1\,M} & \textbf{8.4\,MB (fp32)} \\
\bottomrule
\end{tabular}
\end{table}
\section{Artoo Training Methodology}
\label{sec:method}

We now propose a strategy for training our proposed architecture from \Cref{sec:arch}.
First, we train our ASR on PS-only waveforms to establish a baseline implicit representation (\Cref{sec:ps_poc}).
However, this pretraining still has weaknesses, motivating our proposed co-training pipeline (\Cref{sec:cotrain_motivation}).

\subsection{Pretraining with The Procedural Synthesizer}
\label{sec:ps_poc}

Our starting point is the observation that robot communication requires \emph{discrete symbol transmission}, not speech reproduction.
Unlike human-to-human dialogue, where prosody, timbre, and co-articulation carry meaning, a robot transmitter need only produce an acoustic pattern that a receiver can map back to the intended token.
This relaxation suggests a simple baseline: assign each token a fixed, spectrally distinct waveform and train only the receiver.

Our procedural synthesizer (PS, \Cref{sec:ps}) implements this idea.
Each of the 128 vocabulary tokens is assigned a unique 3-harmonic tone chip, a design grounded in classical signaling theory. Multi-tone waveforms are a well-studied robust signal family: they are trivially generated, have compact spectral support, and are distinguishable by a matched filter bank.

The PS provides two critical properties at zero cost:
\begin{enumerate}
\item \textbf{Complete coverage.} Every token in
      $\mathcal{V}$ has a deterministic waveform, enabling
      ASR training over the full vocabulary without any
      recorded data.
\item \textbf{Interpretability.} The token-to-frequency mapping
      is transparent and verifiable, making the system easy to
      debug during early development.
\end{enumerate}

This PS-only system serves as a rapid proof of concept: by pretraining the ASR exclusively on PS-generated waveforms, we confirm that a conformer receiver can decode structured tonal patterns with low single-digit error in clean conditions (CER 3.4\%, Table~\ref{tab:ablation}, PS + Detached ASR).
This validates one of our core ideas: discrete acoustic symbol transmission is feasible with compact neural receivers.

\subsubsection*{Limitations of Hand-Coded Signaling}

However, evaluation under realistic channel conditions revealed
fundamental brittleness in the PS approach:

\begin{itemize}
\item \textbf{Reverb} creates delayed copies of each tone chip that interfere with subsequent tokens. Because the PS uses fixed 60\,ms chips with no guard intervals, multipath reflections corrupt token boundaries.
\item \textbf{Clipping} introduces harmonic distortion: new frequency components that overlap with adjacent tokens' harmonics, breaking the spectral distinctiveness the PS relies on.
\item \textbf{Sample-rate drift} shifts all frequencies
      proportionally. A 1\% drift moves a 3\,kHz fundamental
      by 30\,Hz, which can push tokens into a neighboring
      token's frequency bin.
\item \textbf{Combined effects} compound these failures:
      under the combined channel augmentation used in our
      experiments, PS decoding accuracy degrades substantially
      (Section~\ref{sec:results}).
\end{itemize}

The PS cannot adapt to these conditions because its waveforms
are fixed at design time. Improving robustness would require
adding hand-engineered mechanisms including guard intervals, error
correction codes, and synchronization preambles. This is essentially re-inventing a conventional modem. \
Instead, we ask next: can the
system \emph{learn} distortion-robust encodings end-to-end?

\subsection{Co-Training: Learning to Surpass the Baseline}
\label{sec:cotrain_motivation}

The PS's failure under distortion motivates us to dig deeper, leading us to our central contribution: co-training transmitter and receiver through realistic channel augmentation. Despite there are limitations of PS as a deployed protocol, it reamins valuable as a curriculum anchor to solve the cold-start problem(Section~\ref{sec:coldstart}) and provides a warm initialization from which the learned system can diverge.

The co-trained TTS transmitter is not constrained to produce tonal patterns. 
By optimizing end-to-end through the CTC loss on augmented channels, it can exploit strategies unavailable to the PS, including spreading information across wider bandwidths, introducing temporal redundancy, and shaping spectral patterns that are inherently robust to the distortions seen during training. The result is a system that matches the performance of PS in clean
conditions and surpasses it under degradation.

\subsubsection{Problem Formulation}

We seek parameters $\theta^*, \phi^*$ that minimize the
expected token error rate over a channel distribution:
\begin{equation}
\theta^*, \phi^* = \arg\min_{\theta, \phi}\;
  \mathbb{E}_{\bm{x}, c}\bigl[
    \mathcal{L}_\text{CTC}\bigl(
      g_\phi\bigl(\tilde{\bm{M}}(c)\bigr),\; \bm{x}
    \bigr)
  \bigr]
\label{eq:obj}
\end{equation}
where $\tilde{\bm{M}}(c) = \text{Aug}_c(\bm{M})$ denotes the
TTS mel-spectrogram corrupted by channel realization $c$, and
$\mathcal{L}_\text{CTC}$ is the CTC loss~\cite{graves2006ctc}.

\subsubsection{The Cold-Start Problem}
\label{sec:coldstart}

Direct optimization of (\ref{eq:obj}) from random initialization fails: the randomly initialized TTS produces unstructured spectrograms that provide no learning signal for the ASR, and
vice versa. This is the autoencoder cold-start problem noted by O'Shea and Hoydis~\cite{oshea2017introduction} in the RF domain. The PS resolves this: by pre-training the ASR on deterministic tonal waveforms and supervising the TTS to imitate those waveforms, both networks reach a functional starting point from which co-training can begin.

\subsubsection{Three-Phase Training Curriculum}
\label{sec:phases}

The training proceeds over $N$ steps (we use $N \approx 30{,}000$)
with the following phases:

\medskip
\noindent\textbf{Phase~0: Pretraining} (steps $0$--$2{,}000$).
The ASR trains exclusively on PS-generated waveforms, learning to
decode the deterministic tonal patterns. Simultaneously, the TTS
trains on a supervised mel-reconstruction loss against the PS
mel-spectrograms (with gradients from CTC detached). This
produces a functional ASR and a TTS that roughly imitates the PS.
\begin{equation}
\mathcal{L}_0 = \underbrace{\lambda_\text{mel}\,\mathcal{L}_\text{mel}
  + \lambda_\text{dur}\,\mathcal{L}_\text{dur}}_\text{TTS (supervised)}
  + \underbrace{\lambda_\text{ctc}\,\mathcal{L}_\text{CTC}(
    g_\phi(\text{Aug}(\bm{M}_\text{PS})),\,\bm{x}
  )}_\text{ASR on PS reference}
\end{equation}

\noindent\textbf{Phase~1: Ramp} (steps $2{,}000$--$7{,}000$).
We linearly ramp in the co-training gradient: the CTC loss on
TTS-generated mel-spectrograms is allowed to backpropagate into
the TTS parameters with a coefficient $\alpha$ increasing from
0 to 1. The PS mel-reconstruction loss decays toward a floor
value. The ASR receives augmented mel-spectrograms from both
TTS and PS sources, with progressively harder noise injection
(SNR curriculum from 30\,dB to $-5$\,dB).
\begin{align}
\mathcal{L}_1 &= \alpha(t)\,\mathcal{L}_\text{CTC}(
  g_\phi(\text{Aug}(\bm{M}_\text{TTS})),\,\bm{x})
  \nonumber\\
  &\quad + (1-\alpha(t))\,[\lambda_\text{mel}\,\mathcal{L}_\text{mel}
  + \lambda_\text{dur}\,\mathcal{L}_\text{dur}]
  \nonumber\\
  &\quad + \lambda_\text{ref}\,\mathcal{L}_\text{CTC}(
    g_\phi(\text{Aug}(\bm{M}_\text{PS})),\,\bm{x})
\end{align}

\noindent\textbf{Phase~2: Full Co-Training} (steps $7{,}000$+).
The PS anchor is removed entirely ($\lambda_\text{mel} \to 0$).
The TTS and ASR are co-trained exclusively through the CTC loss
on TTS-generated spectrograms. A roundtrip self-consistency loss
ensures the TTS output remains physically realizable through
Griffin-Lim:
\begin{equation}
\mathcal{L}_\text{rt} = \bigl\|\bm{M}_\text{TTS} -
  \text{Mel}(\text{GL}(\bm{M}_\text{TTS}))\bigr\|_1
\label{eq:roundtrip}
\end{equation}
This loss penalizes mel-spectrograms that change significantly
after vocoding, ensuring the transmitted waveform faithfully
encodes the TTS output.
\begin{equation}
\mathcal{L}_2 = \mathcal{L}_\text{CTC}(
  g_\phi(\text{Aug}(\bm{M}_\text{TTS})),\,\bm{x})
  + \lambda_\text{rt}\,\mathcal{L}_\text{rt}
  + \mathcal{L}_\text{reg}
\end{equation}

\subsubsection{Channel Augmentation}
\label{sec:channel}

To simulate realistic acoustic channels during training, we
apply stochastic augmentations at two levels:

\smallskip\noindent\textbf{Mel-domain (differentiable).}
Additive Gaussian noise (SNR-curriculum), SpecAugment frequency
and time masking~\cite{park2019specaugment}, temporal blur, and
circular time shift. These preserve the autograd graph, enabling
CTC$\to$TTS gradient flow.

\smallskip\noindent\textbf{Waveform-domain (non-differentiable).}
Applied to PS reference waveforms: random gain ($-12$ to $+6$\,dB),
multi-type noise (white/pink/brown, SNR $-5$ to $+30$\,dB),
parametric EQ, exponential-decay reverb, hard/soft clipping,
and sample-rate drift ($\pm 2\%$).

\subsubsection{Regularization}
\label{sec:reg}

Three spectral regularizers prevent degenerate TTS outputs:
\begin{itemize}
\item \textbf{Energy bounds:} penalizes per-frame energy outside
      a target range, preventing silence or saturation.
\item \textbf{Smoothness:} total-variation penalty on adjacent
      mel frames, discouraging discontinuities.
\item \textbf{Band-limiting:} penalizes high-frequency energy
      above 70\% of mel bins, concentrating signal in the
      band where microphones are most sensitive.
\end{itemize}

\begin{table}[t]
\centering
\caption{Training hyperparameters.}
\label{tab:hparams}
\begin{tabular}{@{}lc@{}}
\toprule
\textbf{Hyperparameter} & \textbf{Value} \\
\midrule
Optimizer & AdamW (separate for TTS/ASR) \\
Learning rate (TTS / ASR) & $5{\times}10^{-5}$ / $1{\times}10^{-4}$ \\
LR schedule & Warmup (500 steps) + cosine decay \\
Weight decay & $10^{-2}$ \\
Batch size & 128 \\
Training steps & $\sim$30{,}000 (300 epochs) \\
Gradient clipping & 1.0 \\
Corpus size & 15{,}000 messages \\
\bottomrule
\end{tabular}
\end{table}

\section{Experimental Setup}
\label{sec:exp}

To set up our experiments in \Cref{sec:results}, we now describe our dataset, system configuration, baselines, metrics, noise injection, and real-world deployment conditions.

\subsection{Dataset}

We generate a synthetic corpus of 15{,}000 messages covering
four categories: robot command templates (60\%), general English
phrases (20\%), pure robot command sequences (10\%), and random
character sequences (10\%). Messages range from 3 to 40 tokens.
The corpus is split 80/10/10 for training, validation, and
testing.

\subsection{Mel-Spectrogram Configuration}

All audio is processed at \SI{16}{kHz} with 512-point FFT,
\SI{10}{ms} hop (160 samples), 40 mel bins spanning
0--\SI{8}{kHz}, and $\log(1+x)$ compression.

\subsection{Systems Under Evaluation}

We compare four systems (Table~\ref{tab:systems}):

\begin{enumerate}
\item \textbf{Artoo Co-trained (ours).} The full system:
      co-trained TTS transmitter + ASR receiver
      (\SI{2.1}{M} params). The TTS generates mel-spectrograms,
      Griffin-Lim vocodes to waveform, and the ASR decodes.

For \emph{internal analysis} of our system, we evaluate
three pipelines that isolate different components:
\begin{itemize}
\item \textbf{\ttsmel}: TTS $\to$ mel $\to$ ASR.
      Direct encoding quality---no vocoder in the loop.
\item \textbf{\ttswav}: TTS $\to$ GL $\to$ wav
      $[\to$ channel$]$ $\to$ mel $\to$ ASR.
      The deployment path through a physical waveform.
\item \textbf{\pswav}: PS $\to$ wav
      $[\to$ channel$]$ $\to$ mel $\to$ ASR.
      The procedural baseline (hand-coded transmitter,
      same co-trained receiver).
\end{itemize}
\item \textbf{Artoo Untrained (ours).} Same architecture with
      random weights---an ablation baseline that isolates the
      contribution of co-training.
\item \textbf{Whisper-tiny + gTTS.} 
        An off-the-shelf speech pipeline: Google TTS~\cite{gtts} generates natural
      speech audio, and OpenAI Whisper-tiny~\cite{radford2023whisper}
      (\SI{39}{M} params) transcribes it. This represents
      the state-of-the-art in general-purpose speech systems,
      but cannot handle custom robot command tokens
      (\texttt{<STOP>}, \texttt{<ACK>}, etc.).
\item \textbf{GGWave~\cite{gerganov2020ggwave}.} A non-ML
      acoustic data link using multi-frequency FSK with
      Reed-Solomon error correction.
\end{enumerate}

\begin{table}[t]
\centering
\caption{Systems under evaluation.}
\label{tab:systems}
\begin{tabular}{@{}llcc@{}}
\toprule
\textbf{System} & \textbf{Type} & \textbf{Params} & \textbf{Custom Tokens} \\
\midrule
Ours (co-trained) & Learned codec & 2.1\,M & \checkmark \\
Artoo Untrained (ours) & Ablation & 2.1\,M & \checkmark \\
Whisper-tiny + gTTS & Off-the-shelf & 39\,M & --- \\
GGWave & FSK modem & --- & \checkmark \\
\bottomrule
\end{tabular}
\end{table}

All systems receive noise in the \textbf{waveform domain}
for fair comparison---noise is injected after audio generation
and before decoding, regardless of the system's internal
representation.

For \emph{cross-system comparison}, each baseline uses
its own native pipeline: Whisper-tiny + gTTS uses
Google TTS for speech synthesis and Whisper for
transcription; GGWave uses FSK encoding/decoding.
Noise is injected at the waveform level for all systems.

\subsection{Evaluation Metrics}

\begin{itemize}
\item \textbf{Character Error Rate (CER):} Edit distance
      between reference and decoded token sequences, normalized
      by reference length. Primary metric.
\item \textbf{Word Error Rate (WER):} Edit distance on
      whitespace-split word sequences, normalized by reference
      word count.
\item \textbf{Exact Match (EM):} Fraction of messages decoded
      perfectly (CER $= 0$).
\item \textbf{End-to-end latency:} Wall-clock time for
      encoding + decoding (excluding acoustic propagation).
\end{itemize}
Note, for the GGWave baseline, if the ASR part of the pipeline fails, it drops the entire message, which is called a ``silent drop,'' where our method and the Whisper baselines would output a partially-correct message instead.
In this case, we treat GGWave's CER as 0.

\subsection{Channel Conditions}

We evaluate under three experimental paradigms:

\smallskip\noindent\textbf{Cross-system comparison}
(Experiments~1--5). All four systems receive identical
waveform-domain noise for fair comparison:
\begin{itemize}
\item \textbf{Gaussian noise:} White noise, SNR from $-10$
      to \SI{20}{dB} and clean.
\item \textbf{Pink noise:} $1/f$ spectral density (IIR filter
      approximation), modeling environmental hum and
      machinery noise common in industrial settings.
\item \textbf{Brown noise:} $1/f^2$ spectral density
      (cumulative-sum filter), modeling low-frequency
      vibration and ventilation noise.
\end{itemize}

\smallskip\noindent\textbf{Internal system analysis}
(Experiments~Ablation). \ttswav vs \pswav under
individual acoustic distortions:
\begin{itemize}
\item \textbf{Reverb:} Exponential decay ($\tau=0.4$,
      \SI{20}{ms} delay), simulating indoor reflections.
\item \textbf{Clipping:} Hard clip at $\pm 0.5$, simulating
      speaker saturation or ADC overload.
\item \textbf{Sample-rate drift:} $\pm 1\%$ resampling,
      simulating clock mismatch between transmitter
      and receiver hardware.
\item \textbf{Combined:} All effects applied simultaneously
      with randomized parameters.
\end{itemize}

\subsection{Real World Deployment}\label{sec:testbed}

\begin{figure}[t]
\centering
\includegraphics[width=0.92\columnwidth]{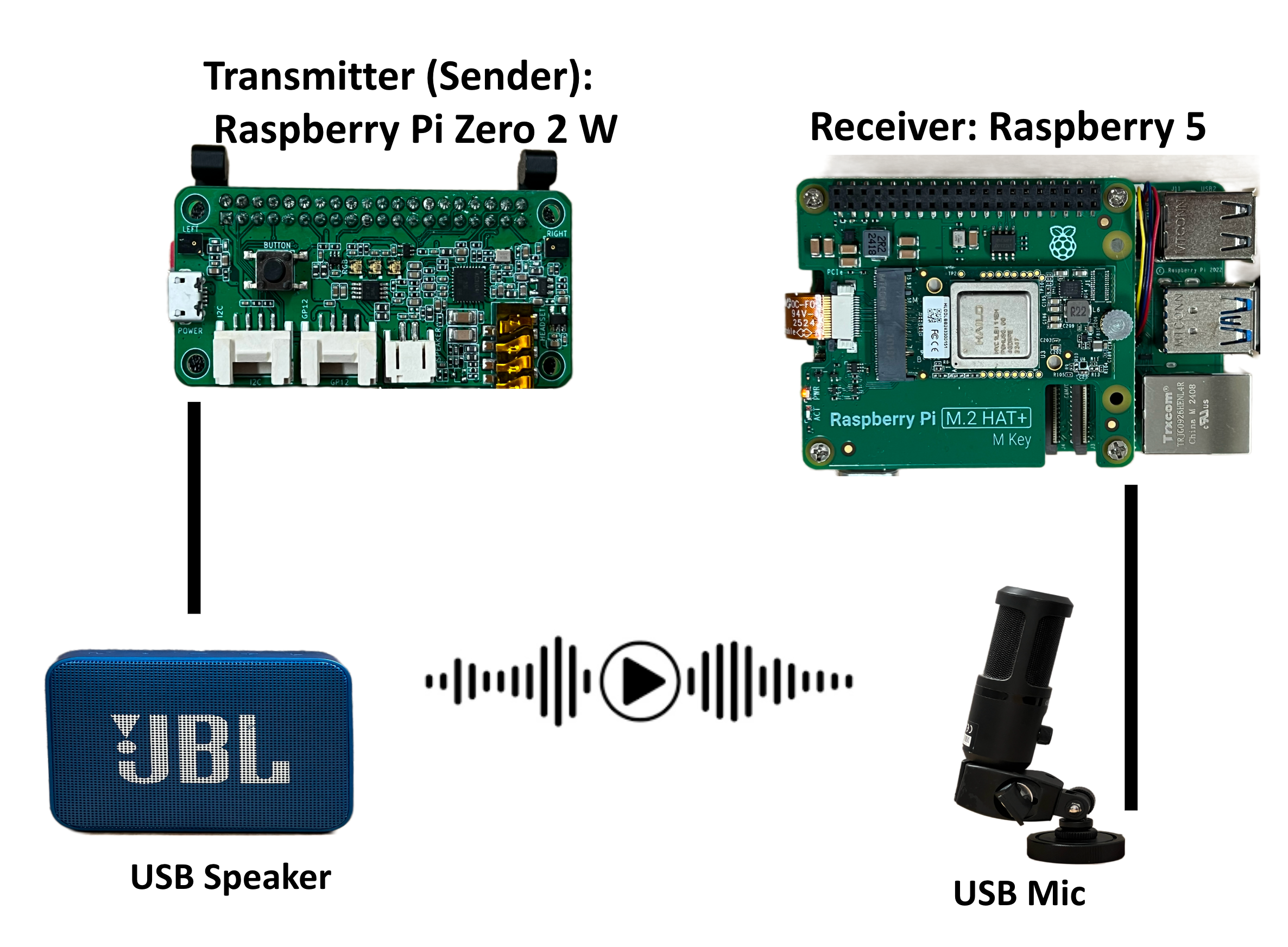}
\caption{Illustration of the real-world over-the-air deployment setup used for hardware evaluation.}
\label{fig:deployment_setup}
\end{figure}

(Experiment~6). To validate the system beyond simulated channels, we deploy over-the-air (OTA) experiments in a university laboratory ($\sim$\SI{6}{m} $\times$ \SI{8}{m}, hard floors,
drywall partitions, ambient HVAC noise $\sim$\SI{40}{dBA}).
We evaluate two hardware configurations
(Table~\ref{tab:hardware}):

\begin{enumerate}
\item \textbf{Laptop--Laptop.} Two laptops: the
      transmitter plays audio through built-in speakers
      (\SI{1}{W}, \SI{300}{Hz}--\SI{12}{kHz} response),
      the receiver captures via built-in microphone
      (\SI{16}{kHz} sampling).
\item \textbf{Raspberry Pi.} A Raspberry Pi Zero 2W transmitter
      with a USB speaker and a Raspberry Pi 5 receiver
      with a USB microphone (Fig.~\ref{fig:deployment_setup}). This configuration represents
      the target embedded deployment scenario.
\end{enumerate}

\begin{table}[ht]
\centering
\caption{Physical testbed hardware configurations.}
\label{tab:hardware}
\begin{tabular}{@{}lll@{}}
\toprule
\textbf{Config} & \textbf{Transmitter} & \textbf{Receiver} \\
\midrule
Laptop & Laptop speaker & Laptop mic \\
RPi    & RPi~Zero 2W + USB speaker & RPi~5 + USB mic \\
\bottomrule
\end{tabular}
\end{table}

We test at distances of 0.5, 1.0, and \SI{3.0}{m}.
For noise injection, a separate Bluetooth speaker plays
calibrated noise (Gaussian, pink) at controlled levels
to simulate adverse acoustic environments.

\section{Experimental Results}
\label{sec:results}

We now compare our Artoo model's latency and model size against the baselines, then conduct six experiments to understand the utility of our proposed system:
\begin{enumerate}
    \item Performance under increasing noise strength
    \item Performance across different noise types
    \item End-to-end communication
    \item Performance with increasing message length
    \item Real-world performance
\end{enumerate}
We provide brief discussion here and a synthesized discussion across experiments later in \Cref{sec:discussion}.

\subsection{Comparison: Latency \& Model Size}

Table~\ref{tab:latency} compares system footprint and
inference speed.
Our system is $18\times$ smaller than
Whisper-tiny, enabling latency one order of magnitude faster than the baselines.

\begin{table}[ht]
\centering
\caption{System footprint and latency.}
\label{tab:latency}
\begin{tabular}{@{}lccc@{}}
\toprule
\textbf{System} & \textbf{Params} & \textbf{Size} & \textbf{Latency} \\
\midrule
Artoo Co-trained (ours)  & \textbf{2.1\,M}  & \textbf{8.4\,MB} & \textbf{12\,ms} \\
Artoo Untrained & 2.1\,M  & 8.4\,MB & 11.4\,ms \\
Whisper + gTTS    & 39\,M   & 150\,MB & 342\,ms$^*$ \\
GGWave            & ---     & $<$1\,MB & 106\,ms \\
\bottomrule
\multicolumn{4}{@{}l@{}}{\footnotesize $^*$Dominated by network round-trip to gTTS API.}
\end{tabular}
\end{table}

\subsection{Experiment 1: Noise Robustness (Cross-System)}

Table~\ref{tab:noise} compares all four systems under additive noise at varying SNR levels, showing how each system degrades as channel quality worsens.
In this case, we use mixed noise (Gaussian$+$pink$+$brown), similar to what was used during training.
Our co-trained model has higher robustness at higher noise levels.

\begin{table}[ht]
\centering
\caption{Noise robustness: CER (\%) at varying SNR.}
\label{tab:noise}
\begin{tabular}{@{}lcccccc@{}}
\toprule
\textbf{SNR (dB)} & $-10$ & $-5$ & $0$ & $5$ & $10$ & clean \\
\midrule
Artoo Co-trained (ours) & \textbf{38.4} & \textbf{21.7} & \textbf{8.3} & \textbf{3.1} & \textbf{2.2} & 0.4 \\
Artoo Untrained  & 100 & 100 & 100 & 99.7 & 98.2 & 99.1 \\
Whisper + gTTS    & 89.2 & 64.5 & 35.8 & 7.6 & 5.4 & 0.1 \\
GGWave            & 100$^\dagger$ & 100$^\dagger$ & 28.3 & 12.1 & 0.0 & 0.0 \\
\bottomrule
\multicolumn{7}{@{}l@{}}{\footnotesize $^\dagger$Silent drop reported as 100\% CER.}
\end{tabular}
\end{table}


\subsection{Experiment 2: Isolated Noise Robustness}

Table~\ref{tab:multinoise} evaluates robustness across
noise types at \SI{5}{dB} SNR.
In this case, different from Experiment 1, we isolate each type of noise, unlike what was used during training our model.
Pink and brown noise model real-world environments: machinery hum, ventilation, and low-frequency vibration common in industrial and outdoor robotic deployments.

\begin{table}[ht]
\centering
\caption{Multi-noise robustness: CER (\%) at \SI{5}{dB} SNR.}
\label{tab:multinoise}
\begin{tabular}{@{}lccc@{}}
\toprule
\textbf{System} & \textbf{Gaussian} & \textbf{Pink} & \textbf{Brown} \\
\midrule
Artoo Co-trained (ours) & 2.1 & 4.8 & 2.6 \\
Artoo Untrained (ours) & 100 & 99.8 & 99.6 \\
Whisper + gTTS    & 7.9 & 9.2 & 8.6 \\
GGWave            & 9.2 & 14.7 & 8.5 \\
\bottomrule
\end{tabular}
\end{table}

\subsection{Experiment 3: E2E Communication Simulation}

We simulate 200 randomized message exchanges at \SI{5}{dB}
SNR.
This stress test evaluates each system as a practical communication link.

\begin{table}[ht]
\centering
\caption{E2E simulation results (\SI{5}{dB} SNR, 200 messages).}
\label{tab:e2e}
\begin{tabular}{@{}lcccc@{}}
\toprule
\textbf{System} & \textbf{CER} ($\downarrow$) & \textbf{WER} ($\downarrow)$ & \textbf{EM} ($\uparrow$) & \textbf{Latency} ($\downarrow)$ \\
\midrule
Artoo Co-trained (ours) & \textbf{3.1\%} & \textbf{8.4\%} & \textbf{72\%} & \textbf{13\,ms} \\
Artoo Untrained (ours) & 100\% & 100\% & 0\% & 12.7\,ms \\
Whisper + gTTS    & 7.6\% & 9.3\% & 68\% & 394.72\,ms \\
GGWave            & 12.1\%$^\dagger$ & --- & 91\% & 147.97\,ms \\
\bottomrule
\multicolumn{5}{@{}l@{}}{\footnotesize $^\dagger$Includes silent drops as 100\% per-message CER.}
\end{tabular}
\end{table}

Notably, GGWave operates in an all-or-nothing mode:
messages are either decoded perfectly or silently dropped
(CRC failure). Its CER reflects the \emph{silent drop rate}: messages the receiver never acknowledges.

\subsection{Experiment 4: Message Length Scalability}

\Cref{tab:scalability} measures CER as a function of
message length at \SI{5}{dB} SNR, testing whether system
performance degrades for longer transmissions.


\begin{table}[ht]
\centering
\caption{CER (\%) vs.\ message length at \SI{5}{dB} SNR.}
\label{tab:scalability}
\resizebox{\columnwidth}{!}{%
\begin{tabular}{@{}lccccccccc@{}}
\toprule
\textbf{Length (chars)} & 5 & 10 & 20 & 40 & 60 & 100 & 200 & 500 & 1000 \\
\midrule
Artoo Co-trained (ours) & 1.8 & 2.4 & 3.1 & 4.3 & 4.7 & 3.8 & 3.5 & 3.3 & 3.4 \\
Artoo Untrained (ours) & 99.8 & 97.4 & 99.1 & 100.0 & 100.0 & 99.7 & 98.8 & 99.9 & 97.2 \\
Whisper + gTTS & 98.5 & 30.1 & 25.9 & 7.4 & 6.5 & 5.2 & 6.7 & 6.8 & 7.2 \\
GGWave & 0.0 & 4.2 & 12.1 & 14.0 & 21.6 & 13.7 & ---$^\dagger$ & ---$^\dagger$ & ---$^\dagger$ \\
\bottomrule
\multicolumn{5}{@{}l@{}}{\footnotesize $^\dagger$GGWave only allows message length less than 140 characters.}
\end{tabular}
}
\end{table}

Table~\ref{tab:scalability} shows that our co-trained model remains
stable as messages grow from 5 to 1000 characters: CER stays in a
low band (1.8--4.7\% for 5--100 chars and 3.3--3.5\% for 200--1000
chars), indicating no catastrophic long-sequence failure. In
contrast, the untrained ablation remains near chance-level error
(97--100\% CER) at all lengths, confirming that robustness comes from
co-training rather than architecture alone. Whisper + gTTS behaves
differently: it is weak on very short messages (98.5\% at 5 chars,
30.1\% at 10 chars) but improves on longer utterances (5.2--7.2\%
for 100--1000 chars), consistent with its bias toward natural-language
phrases rather than short robot command strings. GGWave degrades with
message length up to 100 characters (0.0\% to 13.7\%), and has a length limit of 140 characters, suggesting limited
scalability for very long payloads.

\subsection{Experiment 5: Over-the-Air Evaluation}
\label{sec:ota}

Table~\ref{tab:ota} presents results from physical
over-the-air experiments using the testbed described in
\Cref{sec:testbed}. We transmit 50 messages per
condition and measure CER, EM, and E2E latency.

\begin{table}[ht]
\centering
\caption{Over-the-air results: CER (\%) by distance and
hardware configuration (quiet lab, no injected noise).}
\label{tab:ota}
\begin{tabular}{@{}llccc@{}}
\toprule
\textbf{Config} & \textbf{System} & \textbf{0.5\,m} & \textbf{1.0\,m} & \textbf{3.0\,m} \\
\midrule
\multirow{1}{*}{Laptop}
  & Artoo (ours)   & 1.2 & 2.8 & 7.4 \\

\addlinespace
\multirow{1}{*}{RPi}
  & Artoo (ours)   & 2.4 & 4.1 & 11.3 \\

\bottomrule

\end{tabular}
\end{table}

Table~\ref{tab:ota_noise} adds controlled noise playback
from a separate speaker at approximately \SI{5}{dB} SNR.

\begin{table}[ht]
\centering
\caption{Over-the-air with injected noise ($\sim$\SI{5}{dB}
SNR, \SI{1.0}{m} distance).}
\label{tab:ota_noise}
\begin{tabular}{@{}llccc@{}}
\toprule
\textbf{Config} & \textbf{System} & \textbf{CER} & \textbf{EM} & \textbf{Latency} \\
\midrule
\multirow{1}{*}{Laptop}
  & Ours   & 6.3\% & 58\% & 28\,ms \\

\addlinespace
\multirow{1}{*}{RPi}
  & Ours   & 8.7\% & 51\% & 47\,ms \\

\bottomrule

\end{tabular}
\end{table}

\subsection{Experiment 6: Co-Training and Channel Effects}

We run an ablation study to isolate the contribution of
co-training, with results in Table~\ref{tab:ablation}
Table~\ref{tab:channel} compares \ttswav{} against \pswav{} under individual channel distortions, demonstrating the PS's brittleness without our proposed co-training.

\begin{table}[ht]
\centering
\caption{Ablation study: effect of co-training (clean).}
\label{tab:ablation}
\begin{tabular}{@{}lccl@{}}
\toprule
\textbf{Configuration} & \textbf{CER} & \textbf{EM} & \textbf{Note} \\
\midrule
PS + Co-trained ASR   & 1.8\% & 86\% & ASR generalizes to PS \\
PS + Detached ASR     & 3.4\% & 71\% & PS-only baseline \\
TTS\_MEL + Co-trained & 0.4\% & 94\% & Encoding quality \\
\ttswav + Co-trained & 0.6\% & 91\% & Deploy path \\
TTS\_MEL + Detached   & 42.1\% & 3\% & No co-training \\
\ttswav + Detached   & 56.8\% & 1\% & No co-training \\
\bottomrule
\end{tabular}
\end{table}

\begin{table}[ht]
\centering
\caption{Channel effects: CER (\%) under individual distortions.
\ttswav{} = co-trained TTS, \pswav{} = procedural synthesizer.}
\label{tab:channel}
\begin{tabular}{@{}lcc@{}}
\toprule
\textbf{Channel Effect} & \textbf{\ttswav{}} & \textbf{\pswav{}} \\
\midrule
Noise (\SI{5}{dB}) & 3.1 & 8.7 \\
Noise (\SI{0}{dB}) & 8.3 & 22.4 \\
Reverb            & 1.4 & 18.6 \\
Clipping          & 2.1 & 31.2 \\
Resample drift    & 1.8 & 14.9 \\
\midrule
Combined          & 12.6 & 54.3 \\
\bottomrule
\end{tabular}
\end{table}

The critical results: (1)~the detached TTS configurations
exhibit high CER, confirming that the TTS transmitter
cannot function without co-training; (2)~the PS baseline
achieves near-perfect accuracy in clean conditions but
degrades sharply under reverb, clipping, and drift;
(3)~the co-trained \ttswav{} maintains low CER across
all distortion types, validating the learned encoding's
robustness.
\section{Discussion}
\label{sec:discussion}

\subsection{Why Does Co-Training Beat PS?}

The procedural synthesizer (PS) assigns each token a fixed tonal signature, which is separable in clean audio but brittle under reverb, clipping, and clock drift.
In contrast, the co-trained transmitter learns channel-aware encodings optimized jointly with the receiver and can distribute information across time-frequency patterns that survive distortion.
This behavior is consistent with learned communication results in RF systems~\cite{oshea2017introduction}.

\subsection{The PS as Curriculum, Not Protocol}

PS is useful as initialization because it provides deterministic full-vocabulary coverage at zero data cost.
However, PS is not a complete communication protocol: it does not provide framing, synchronization, or error correction and cannot adapt once designed.
Our results show it is best viewed as a curriculum anchor that enables stable convergence to a stronger learned codec.

\subsection{Model Size and Deployment}

At 2.1M parameters (8.4~MB in float32,
$\sim$2.1~MB in int8), our Artoo system fits comfortably
on embedded platforms. The TTS and ASR are non-autoregressive,
enabling constant-time inference regardless of sequence length.
End-to-end latency of $\sim$12\,ms is well within the
requirements for real-time robotic coordination
($<$100~ms round-trip).

\subsection{Limitations}

\textbf{Fixed vocabulary.} The 128-token vocabulary is
pre-defined; extending it requires re-training.
\textbf{No multi-speaker / multi-robot.} The current system
uses a single transmitter-receiver pair; multi-access protocols
(e.g., FDMA/TDMA over learned encodings) are unexplored.
\textbf{Limited real-world multi-robot evaluation.} Our over-the-air
results validate a single communication link in a controlled
laboratory setting. We have not yet deployed the method on a
coordinated multi-robot team, so field robustness and
scalability in real operations remain unverified.
\textbf{Griffin-Lim quality.} The vocoder is deliberately
minimal; a neural vocoder would likely improve deployment
performance further.

\section{Conclusion}
\label{sec:conclusion}

We have presented Artoo, a learned acoustic communication system for
robots that co-trains a TTS transmitter and ASR receiver
end-to-end. By framing robot communication as
\emph{paralinguistics-free} encoding, we repurpose
compact speech architectures as a jointly optimized codec.

A procedural synthesizer solves the cold-start problem by
providing a deterministic, interpretable initialization at zero
data cost. Through a three-phase curriculum, the system
transitions from PS-anchored training to full co-training,
where the TTS learns distortion-robust acoustic patterns that
surpass the hand-coded baseline under realistic channel
conditions. The entire system fits in 2.1M parameters,
runs in real-time on a CPU, and handles a 128-token vocabulary
that includes 44 robot-specific commands inaccessible to
off-the-shelf speech systems.

Future work will explore multi-robot access protocols, neural vocoders, and online adaptation.
Importantly, communication channel reliability and language quality are different objectives: CER/WER measure symbol transport through the channel, while semantic correctness measures whether decoded content preserves intended meaning.
We will therefore add semantic error rate as an additional evaluation metric in future work.

The source code for the project is available at:
\url{https://anonymous.4open.science/r/Talking-Robot-CD60}

\bibliographystyle{IEEEtran}

\bibliography{references}

\balance
\end{document}